%% file: bmvc_final.tex
\PassOptionsToPackage{table}{xcolor}
\documentclass{bmvc2k}

\input{setup/package} 
\input{setup/macros} 
\input{setup/symbols} 
\input{setup/graphicspath} 
\usepackage{wrapfig}
\usepackage{kotex}

\title{HERO-VQL: Hierarchical, Egocentric \\ and Robust Visual Query Localization}

\addauthor{Joohyun Chang$^*$}{joohyun7u@khu.ac.kr}{1} 
\addauthor{Soyeon Hong$^*$}{soyeonhong@khu.ac.kr}{1}
\addauthor{Hyogun Lee$^{*}$}{gunsbrother@khu.ac.kr}{1}
\addauthor{Seong Jong Ha}{sj.ha1@cj.net}{2}
\addauthor{Dongho Lee}{dongho.lee.14@cj.net}{2}
\addauthor{Seong Tae Kim$^\dagger$}{stkim@khu.ac.kr}{1}
\addauthor{Jinwoo Choi$^\dagger$}{jinwoochoi@khu.ac.kr}{1}

\addinstitution{
 Kyung Hee University \\
 Republic of Korea
}
\addinstitution{
 AI R\&D Division, CJ Group \\
 Republic of Korea \\
 \vspace{0.3em}
 \hspace{-0.25cm} $^*$Equal contribution \\
 \hspace{-0.25cm} $^\dagger$Corresponding authors \\
}

\runninghead{Chang et al.}{HERO-VQL: Hierarchical, Egocentric and Robust VQL}

\def\eg{\emph{e.g}\bmvaOneDot}

\begin{document}
\maketitle

\input{0_abstract}

\input{1_introduction}

\input{2_prior}
\input{4_method}
\input{5_result}
\input{6_conclusion}

\section*{Acknowledgements}

\noindent
This work was supported in part by AI R\&D Division, CJ Group, the Institute of Information \& Communications Technology Planning \& Evaluation (IITP) grant funded by the Korea Government (MSIT) under grant RS-2024-00353131 (20\%), RS-2021-0-02068 (Artificial Intelligence Innovation Hub, 20\%), and RS-2022-00155911 (Artificial Intelligence Convergence Innovation Human Resources Development (Kyung Hee University, 20\%)),
Additionally, it was supported by the Institute of Information \& Communications Technology Planning \& Evaluation(IITP)-ITRC(Information Technology Research Center) grant funded by the Korea government (MSIT)(IITP-2025-RS-2023-00259004, 20\%), and the National Research Foundation of Korea(NRF) grant funded by the Korea government(MSIT)(IRIS RS-2025-02216217, 20\%).

\bibliography{main}
\clearpage
\appendix

\section*{Supplementary Material}
\input{50_supp_content}
\end{document}

%% file: setup/package.tex
\usepackage{graphicx}
\usepackage{float}
\usepackage{lscape}                       

\usepackage[lined,ruled,linesnumbered]{algorithm2e}

\usepackage{booktabs}                   
\usepackage{multirow}

\usepackage{paralist}

\usepackage{bm}                          
\usepackage{epsfig}                      
\usepackage{times}
\usepackage{mathptmx}
\usepackage{mathtools}
\usepackage{amssymb,amsmath}   
\usepackage{bbm}
\usepackage{pifont}

\usepackage[normalem]{ulem}  
\usepackage{units}
\usepackage{color}

\usepackage{comment}

\usepackage[capitalize]{cleveref}
\crefname{section}{Sec.}{Secs.}
\Crefname{section}{Section}{Sections}
\Crefname{table}{Table}{Tables}
\crefname{table}{Tab.}{Tabs.}

\usepackage{xspace}
\usepackage{setspace}

%% file: setup/macros.tex
\def\eg{e.g.,~}               
\def\ie{i.e.,~}               

\DeclareMathOperator*{\argmax}{\arg\!\max}

\newlength\paramarginsize
\newlength\figmarginsize
\newlength\secmarginsize
\newlength\figcapmarginsize
\newlength\tabcapmarginsize

\newcommand{\secmargin}{\vspace{\secmarginsize}}
\newcommand{\paramargin}{\vspace{\paramarginsize}}
\newcommand{\figmargin}{\vspace{\figmarginsize}}
\newcommand{\figcapmargin}{\vspace{\figcapmarginsize}}
\newcommand{\tabcapmargin}{\vspace{\tabcapmarginsize}}

\newcommand{\mpage}[2]
{
\begin{minipage}{#1\linewidth}\centering
#2
\end{minipage}
}

\newcommand{\figcaption}[2]
{
\caption{
\textbf{#1.}  
#2            
}
}

\newcommand{\secref}[1]{Section~\ref{sec:#1}}
\newcommand{\figref}[1]{Figure~\ref{fig:#1}} 
\newcommand{\tabref}[1]{Table~\ref{tab:#1}}

\newcommand{\algref}[1]{Algorithm~\ref{#1}}

\long\def\ignorethis#1{}

\newcommand{\tb}[1]{\textbf{#1}}

%% file: setup/symbols.tex
\def\xi{\mathbf{x}_i}

\def\C{\mathbf{C}}
\def\I{\mathbf{I}}

\def\C{\mathbf{C}}

\def\Q{\mathbf{Q}}
\def\W{\mathbf{W}}
\def\K{\mathbf{K}}
\def\R{\mathbb{R}}
\def\S{\mathbf{S}}
\def\U{\mathbf{U}}
\def\V{\mathbf{V}}
\def\Y{\mathbf{Y}}
\def\Vt{\mathbf{V}^\top}

\def\s{\mathbf{s}}

\def\Ivideo{\mathbf{I}_{\text{video}}}

\def\Iquery{\mathbf{I}_{\text{query}}}
\def\Zvideo{\mathbf{Z}_{\text{video}}}
\def\zvideo{\mathbf{z}_{\text{video}}}
\def\Zquery{\mathbf{Z}_{\text{query}}}
\def\tZquery{\mathbf{\tilde{Z}}_{\text{query}}}
\def\Zcls{\mathbf{Z}_{\text{cls}}}

\def\ahigh{\alpha_\text{high}}
\def\amid{\alpha_\text{mid}}

\def\ltoken{L_\text{token}}
\def\lmap{L_\text{map}}

\def\ours{\texttt{HERO-VQL}}

\definecolor{graygray}{HTML}{bbbbbb}
\def\tap{$\text{tAP}_{25}$}
\def\stap{$\text{stAP}_{25}$}

\newcommand{\cmark}{\ding{51}}%
\newcommand{\graycross}{{\color{graygray}\ding{55}}}

%% file: setup/graphicspath.tex
\graphicspath{{figure}, {images}, {example}}

%% file: 0_abstract.tex
\begin{abstract}
In this work, we tackle the egocentric visual query localization (VQL), where a model should localize the query object in a long-form egocentric video. 
Frequent and abrupt viewpoint changes in egocentric videos cause significant object appearance variations and partial occlusions, making it difficult for existing methods to achieve accurate localization.
To tackle these challenges, we introduce Hierarchical, Egocentric and RObust Visual Query Localization (\ours{}), a novel method inspired by human cognitive process in object recognition.
We propose i) Top-down Attention Guidance (TAG) and ii) Egocentric Augmentation based Consistency Training (EgoACT).
Top-down Attention Guidance refines the attention mechanism by leveraging the class token for high-level context and principal component score maps for fine-grained localization.
To enhance learning in diverse and challenging matching scenarios, EgoAug enhances query diversity by replacing the query with a randomly selected corresponding object from ground-truth annotations and simulates extreme viewpoint changes by reordering video frames.
Additionally, CT loss enforces stable object localization across different augmentation scenarios.
Extensive experiments on VQ2D dataset validate that \ours{} effectively handles egocentric challenges, significantly outperforming baselines.

\end{abstract}

%% file: 1_introduction.tex
\section{Introduction}
\label{sec:intro}
Egocentric visual query localization (VQL) is a challenging task that aims to spatio-temporally localize the last occurrence of a visual query object within a long-form egocentric video.
In \figref{teaser} (a), we illustrate the egocentric VQL task in the Ego4D dataset~\cite{grauman2022ego4d}.
A successful egocentric VQL method significantly enhances everyday life through applications such as intelligent AR glasses~\cite{ar-glass} or mobile robot assistants~\cite{5587059}.

Egocentric videos pose significant challenges due to frequent and abrupt viewpoint changes caused by the camera wearer’s movements~\cite{grauman2022ego4d, TREK150iccvw, Damen2018EPICKITCHENS, xu2023my}.
As illustrated in \figref{teaser} (b), third-person videos typically maintain stable object visibility. 
In contrast, egocentric videos often contain objects undergoing substantial appearance changes and occlusions due to abrupt viewpoint changes. 
For example, as shown in \figref{teaser} (c), a banana can appear drastically different depending on the camera angle and sometimes a bottle is only partially visible.

Existing egocentric visual query localization works have shown great progress by using a detector and a tracking model~\cite{tang2023egotrackslongtermegocentricvisual, grauman2022ego4d, mai2023egoloc, xu2022negative, xu2023my} and end-to-end learning spatio-temporal relationships~\cite{jiang2023vqloc}.
Despite the great progress, current VQL methods still struggle to accurately localize objects due to the aforementioned challenges in egocentric videos.

Unlike existing egocentric VQL methods, humans can accurately locate objects even under frequent and abrupt viewpoint changes due to ego-motion.
For precise object localization, humans follow a top-down perceptual process~\cite{Reverse_hierarchy, view_from_top}. 
They first recognize objects at a high-level and then refine their perception by focusing on details.

\input{figure/fig_teaser}

In this work, we introduce Hierarchical, Egocentric and RObust Visual Query Localization ({\ours}), a new egocentric visual query localization method.
First, inspired by the human perception process, we design Top-down Attention Guidance (TAG) to enhance the attention mechanism of our spatial decoder.
We begin by guiding the attention process at a high level, using the class token of the object as a query during attention with video features. 
This allows the model to capture the global object context, helping it recognize the object even when it appears at different scales or angles.
Next, we refine attention by guiding the model to focus on object details using principal component (PC) score maps derived from the query image.
The PC score maps improve the model’s ability to precisely localize objects, even under varying perspectives or partial occlusions.

Second, to enable robust matching between visual query and object instances under challenging egocentric video conditions, we propose Egocentric Augmentation based Consistency Training (EgoACT).
Egocentric Augmentation (EgoAug) simulates challenging real-world ego-motions.
To this end, EgoAug comprises two types of augmentations:
i) replacing the query image with a randomly selected corresponding object from the ground-truth (GT) annotations, allowing the model to learn from a broader range of query-GT pairs, 
ii) Reordering video frames to simulate more abrupt object and camera motion.
With Consistency Training (CT) loss, we penalize the model when the predictions differ across augmented clips encouraging more consistent localization.
Consequently, EgoACT improves the model’s robustness to appearance variations, abrupt motions, and partial occlusions.

To validate the effectiveness of {\ours}, we design a set of controlled experiments on the VQ2D~\cite{grauman2022ego4d} dataset.
\ours{} outperforms existing VQL methods.

In this work, we make the following major contributions:
\vspace{-0.5em}
\begin{itemize}
    \item We propose Top-down Attention Guidance (TAG), a hierarchical attention mechanism to refine the attention of the spatial decoder.
    \vspace{-0.5em}
    \item We introduce Egocentric Augmentation based Consistency Training (EgoACT) to improve robustness against appearance changes, abrupt motion, and partial occlusions.
    
    \vspace{-0.5em}
    \item We validate the effectiveness of \ours{} through extensive experiments, demonstrating strong performance on egocentric video: VQ2D~\cite{grauman2022ego4d}.
\end{itemize}
\paramargin

%% file: figure/fig_teaser.tex
\begin{figure}[t]
    \includegraphics[width=\textwidth]{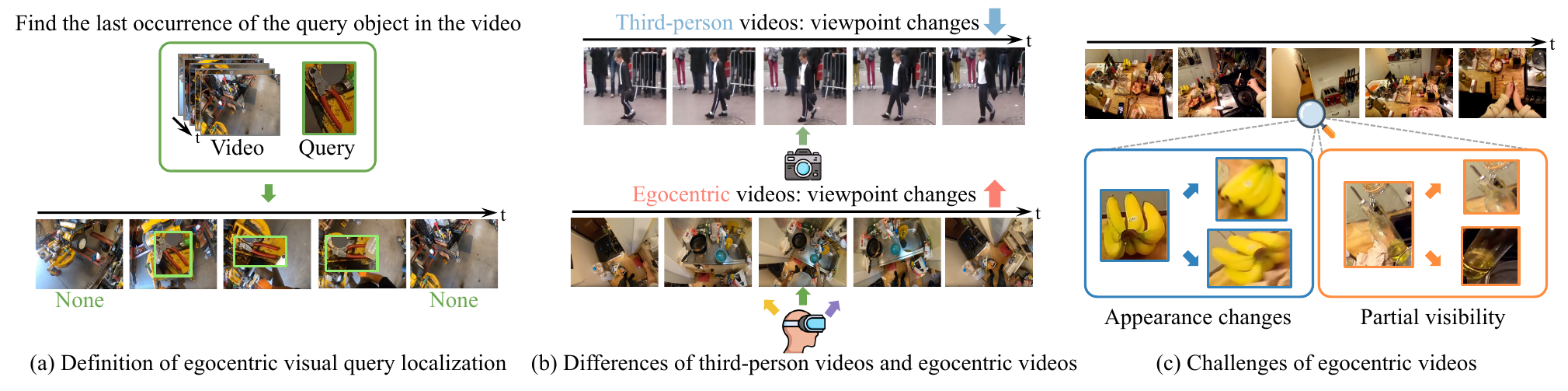}
    \vspace{-1.5em}
    \figcaption{Egocentric visual query localization (VQL)}{
    (a) Given an egocentric video and a query image of an object, the goal is to localize the last occurrence of the query object in the video.
    (b) Unlike third-person videos, egocentric videos undergo abrupt viewpoint changes due to the camera wearer’s movements.
    (c) These viewpoint changes introduce significant challenges in VQL, including variations in object appearance across perspectives and partial visibility when objects move out of the frame.
    For example, the appearance of a banana changes depending on the viewpoint, and a bottle becomes partially visible.
    }
    \label{fig:teaser}
    \vspace{-1em}
\end{figure}

%% file: 2_prior.tex
\section{Related Work}
\label{sec:related}

\paragraph{Egocentric visual query localization.}
With the advancement of egocentric datasets~\cite{grauman2022ego4d, islam2024video, tang2023egotrackslongtermegocentricvisual,grauman2024egoexo4dunderstandingskilledhuman,Damen2018EPICKITCHENS,song2023egod,goletto2024amego,Kurita_2023_ICCV}, there has been significant progress in egocentric visual query localization (VQL), where the query object may come from outside the video.
Key works on egocentric visual query localization include using object detector and tracking-based methods~\cite{xu2023my, xu2022negative, mai2023egoloc, tang2023egotrackslongtermegocentricvisual,huang2023egocentric}, transformer architectures~\cite{jiang2023vqloc} and data augmentations~\cite{xu2023my, xu2022negative}.
Despite these advancements, existing methods still struggle with accurate object localization due to the challenges inherent in egocentric videos, such as frequent and abrupt viewpoint changes, significant object appearance variations, and partial occlusions.
Unlike prior approaches, we tackle these challenges by leveraging learning from diverse matching scenarios and a top-down guided attention mechanism, enabling more robust and accurate localization.%

\paramargin
\paragraph{Learning high-level information for object localization.}
Capturing high-level information about the target object is crucial for effective object localization.
Key approaches include contrastive learning~\cite{chen2020simple,he2020momentum,grill2020bootstrap,caron2020unsupervised,oquab2024dinov}, attention-based representation learning~\cite{grauman2022ego4d,Cui_2022_CVPR,yan2021learning,chen2023seqtrack,jia2023h-detr,song2022transformer,jiang2023vqloc} and using visual reference prompt~\cite{Zhu_2023_CVPR,yang2024samuraiadaptingsegmentmodel,jiang2025sam2motnovelparadigmmultiobject}.
While these methods have shown success, they often overlook valuable \emph{global} information contained in the class token in transformers~\cite{grauman2022ego4d,Cui_2022_CVPR,yan2021learning,chen2023seqtrack,jia2023h-detr,song2022transformer}.
Additionally, contrastive learning methods require large batch sizes~\cite{chen2020simple,he2020momentum,grill2020bootstrap,caron2020unsupervised,oquab2024dinov} and visual reference prompts depend on additional modality inputs — both of which can be computationally impractical for VQL tasks, especially in academic settings.
In contrast to prior approaches, we leverage \emph{global} cues from the class token for guiding the attention process, enabling more robust object localization in challenging egocentric videos.

\paramargin
\paragraph{Using mid-level information for object localization.}
Real-world visual localization tasks often face challenges such as occlusions and appearance variations, making mid-level information crucial for refining object representations and improving robustness.
Existing approaches address these challenges through attention maps~\cite{chen2023category, zhang2018self, li2023locate} and part-based region matching~\cite{cho2015unsupervised, chen2014detectcandetectingrepresenting, Fang_2018_ECCV, li2023locate}, which have demonstrated effectiveness in handling occlusions and object appearance variations.
Unlike prior methods that incorporate mid-level information through detection pipelines or attention mechanisms, we leverage principal component score maps derived from query and video feature. 
This allows \ours{} to capture fine-grained details, making it particularly well-suited for the egocentric VQL task.

%% file: 4_method.tex
\section{\ours{}}
\label{sec:method}
We propose \ours{}, a novel egocentric visual query localization (VQL) method that integrates hierarchical attention mechanisms with training strategies specifically designed to address the challenges of egocentric VQL. 
In this section, we first provide an overview of the proposed method in \secref{overall_architecture}.
Then we provide details of 
i) top-down attention guidance (TAG) in \secref{tag}, ii) egocentric augmentation (EgoAug) in \secref{egoaug}, and consistency training (CT) loss in \secref{ctloss}.


\paramargin
\subsection{Overview}
\label{sec:overall_architecture}
\vspace{-0.5em}
The egocentric VQL task aims to spatio-temporally localize the last occurrence of an object within a long-form egocentric video, given a query image of the object.
We show an overview of the proposed method in \figref{overview} (a).

\paramargin
\paragraph{Formulation.}
Given a RGB video $\Ivideo \in \mathbb{R}^{T \times C \times H \times W}$ with $T$ frames, $C$ channels, and $H \times W$ spatial dimensions, and a query image $\Iquery \in \mathbb{R}^{C \times H \times W}$ as input, the goal is to predict the temporal segment where the object last appears and localize it in each frame of this segment.
Each bounding box is represented as $\mathbf{B}_i = \{ x_i, y_i, w_i, h_i \}$, where $(x_i, y_i)$ denote the top-left coordinates, and $(w_i, h_i)$ represent the width and height of the object in the $i$-th frame.


\input{figure/fig_overview}

\paramargin
\paragraph{Egocentric Augmentation based Consistency Training.}
During training, we enhance the model's robustness to frequent and abrupt viewpoint changes in egocentric videos by using Egocentric Augmentation based Consistency Training (EgoACT), as illustrated in \figref{overview}~(a).
EgoACT consists of two components.
First, Egocentric Augmentation (EgoAug) increases query diversity and simulates abrupt egocentric motion during training.  
EgoAug replaces the query image $\Iquery$ with another instance $\mathbf{I}_{\text{QueryAug}}$ randomly selected from the ground-truth object instances in the video.
Then EgoAug reorders the input video $\mathbf{I}_\text{video}$ to simulate abrupt viewpoint changes, resulting in $\mathbf{I}_\text{MotionAug}$.
Second, Consistency Training (CT) loss enforces consistency between predictions across augmented clips. 
We provide further details on EgoAug and CT loss in \secref{egoaug}, and \secref{ctloss}, respectively.


\paramargin
\paragraph{Encoder.}
We use a pre-trained encoder $f_E(\cdot)$, based on DINOv2~\cite{oquab2024dinov}, to extract features from each frame of the video $\Ivideo$ and the query image $\Iquery$.
Consequently, we obtain video features $\Zvideo \in \mathbb{R}^{T \times M \times D}$ and the query features $\Zquery \in \mathbb{R}^{N \times D}$, where $D$ is the embedding dimension, and $M$ and $N$ are the number of video and query patch tokens, respectively.


\paramargin
\paragraph{Spatial Decoder.} 
The spatial decoder $f_D(\cdot)$ refines object localization between the video and the query image.
We adopt a standard transformer-based architecture~\cite{vaswani2017attention} for the spatial decoder.
Specifically, the video features $\Zvideo$ serve as the query, while the query features $\Zquery$ serves as the key and value. 
To enhance localization, we introduce Top-down Attention Guidance (TAG), inspired by human visual recognition.
A high-level attention guide ($\ahigh$) directs the model to focus on query-relevant video regions, while a mid-level attention guide ($\amid$) refines localization by emphasizing distinctive object parts.
These attention maps help the decoder progressively narrow down the search space and improve localization.
The decoder outputs object-aware features $\Y \in \R^{T \times M \times D}$, capturing interactions between the video and the query.
We provide further details on TAG in \secref{tag}.


\paramargin
\paragraph{Temporal Context Modeling.} 
To model temporal context, we apply a temporal module $f_T(\cdot)$, based on TSM~\cite{lin2019tsm}, to the object-aware features $\Y$ to obtain the spatio-temporal features $\W = f_T(\Y) \in \R^{T \times M \times D}$.
The spatio-temporal features capture inter frame-level relationships and enrich the model’s ability to understand object movements over time.


\paramargin
\paragraph{Prediction Head.} 
Following prior work~\cite{jiang2023vqloc}, \ours{} has a box prediction head and a score prediction head.
These heads take the spatio-temporal features $\W$ as an input and predict a bounding box $\mathbf{B}_i = \{ x_i, y_i, w_i, h_i \}$ and the confidence score $\pi_i$ of $i$-th frame.


\paramargin
\subsection{Top-down Attention Guidance} 
\label{sec:tag}
We propose Top-down Attention Guidance (TAG) to enhance the spatial decoder’s ability to localize objects under egocentric viewpoint changes.
TAG enables the spatial decoder to process the query object in a hierarchical manner:
In the early attention layers, TAG provides high-level guidance for object-level localization, helping the model capture the global information of the query.
In the later attention layers, TAG refines localization with mid-level guidance, focusing on fine-grained object parts for precise matching.

\vspace{-0.5em}
\subsubsection{High-level Attention Guide}
\label{sec:hag}

High-level attention guidance captures the global object context, allowing the model to focus on query-relevant regions in the video.  
Inspired by human perception, where objects are first recognized at a category level before utilizing fine-grained details~\cite{Reverse_hierarchy, view_from_top}, we leverage the class token of the query image as a semantic reference to guide the decoder’s attention.  
As shown in \figref{overview}~(b), we compute attention scores between the class token $\Zcls^{L-1}\in \mathbb{R}^{D}$ extracted from the penultimate layer of the encoder and the patch tokens $\zvideo^{L-1}\in \mathbb{R}^{M \times D}$ of the video frames .
Then we compute the high-level attention guide $\ahigh$ as follows:
\vspace{-0.2em}
\begin{equation} 
\label{eq:hag_mask_eq} 
    \ahigh = \sigma\left( \Zcls^{L-1} \W_{Q}(\zvideo^{L-1} \W_{K})^\top \right) \in \R^{M}, 
\end{equation} 
\vspace{-0.2em}
where $\sigma(\cdot)$ denotes the frame-level mean score subtraction followed by the sigmoid function, $\mathbf{W}_{Q}$ and $\mathbf{W}_{K}$ are the $D \times D$ query and key projection matrices of the penultimate layer of the encoder, respectively.

The spatial decoder adds the high-level attention guide $\ahigh$ to the self-attention map computed between the query and key. 
The resulting attention scores are then passed through a softmax to guide the focus toward video regions that are more relevant to the query object.

\paramargin

\subsubsection{Mid-level Attention Guide}
\label{sec:lag}

Mid-level attention guidance refines localization by directing attention toward specific object parts.  
Once the object is roughly localized, humans refine their perception by focusing on distinctive local features~\cite{Reverse_hierarchy, view_from_top}.  
Following this principle, we enhance localization by directing attention to specific object parts.  
We extract the most discriminative structures of the query object using principal component analysis (PCA) on the query features.

\paramargin
\paragraph{Principal Component.}

We first center the query features $\Zquery$ to obtain $\tZquery$, then we perform a low-rank principal component decomposition: $\tZquery \approx \U\Sigma\Vt_{\tZquery}$. 
We retain the top-$R$ singular vectors, denoted as $\V_{\tZquery} \in \R^{D \times R}$, which capture the most significant variations in the query feature.

\paramargin
\paragraph{PC score map.}
As shown in \figref{overview} (b), we compute the mid-level attention guide $\alpha_\text{mid}$ as follows:
\begin{equation}
\label{eq:lag_score_map_eq}
    \alpha_\text{mid} = \phi(\tZquery \V_{\tZquery}) \in [0, 1)^{N \times R}, \quad \text{where } \phi(x) = 1 - e^{-x^2 / \tau}. 
\end{equation}
Here, $\phi(\cdot)$ is an element-wise activation function that adjusts the influence of negative singular vector directions, $\tau$ is a scaling parameter that controls the sharpness of the calibration curve.
This process generates $R$ score maps $\S$, each highlighting different query characteristics, \eg grip, or base of the pan in \figref{overview}.

\paramargin
\paragraph{Cross-attention in the spatial decoder.}
In the spatial decoder, we incorporate the mid-level attention guide $\alpha_\text{mid}$ into the multi-head cross-attention mechanism by assigning each of the $R$ PC score maps to one of the $R$ attention heads. 
Each head adds the assigned score map to the cross-attention map before applying softmax.
Since each score map encodes distinct object-part information, the heads attend to different object parts, enabling part-specific attention that benefits localization under partial visibility and viewpoint changes.
For more details, please refer to the supplementary material.


\paramargin
\subsection{Egocentric Augmentation} 
\label{sec:egoaug}
To address egocentric VQL challenges, we introduce Egocentric Augmentation (EgoAug), which consists of two augmentations: Visual Query Augmentation (QueryAug) and Ego-motion Augmentation (MotionAug).

\paramargin
\paragraph{QueryAug.} We propose QueryAug, a simple yet effective query augmentation technique designed to leverage the diverse correspondence signals present in ground-truth annotations explicitly.
As illustrated in \figref{overview}~(c), given a query image $\Iquery$ containing an object, we gather all instances of the object in the video, denoted by $\mathbb{I}_{\text{obj}} = \{\I_1, \I_2, \ldots, \I_k\}$, where $k$ is the total number of target object instances.
We then randomly select one instance $\mathbf{I}_\text{QueryAug}$ from $\mathbb{I}_{\text{obj}}$ to replace the query image with a replacement probability $p\in[0,1]$.
By incorporating QueryAug, we encourage the model to learn robust object correspondence across \emph{diverse variations}, such as object pose, motion blur, and partial occlusion induced by ego-motion.

\paramargin
\paragraph{MotionAug.} We propose MotionAug, a data augmentation technique designed to simulate abrupt viewpoint changes and dynamic motion changes in egocentric videos.
As illustrated in \figref{overview}~(c), MotionAug amplifies object motion dynamics by reordering frames within the ground-truth (GT) segment of a video.
This process increases the speed of bounding box movements, inducing more abrupt viewpoint shifts and challenging localization conditions.
Given a RGB video $\Ivideo$, we construct a reordered version $\mathbf{I}_\text{MotionAug}$ as follows.
First, we assign the initial frame of the GT temporal segment as the first frame of $\mathbf{I}_\text{MotionAug}$.
At each step, we select the GT frame that exhibits the largest bounding box displacement relative to the bounding box in the previous frame in the $\mathbf{I}_\text{MotionAug}$.
We repeat this process until all GT-annotated frames are reordered.
To quantify displacement, we compute variations in bounding box width, height, centroid position, and their temporal derivatives.
We put further details in the supplementary material.
By training with MotionAug, \ours{} improves robustness to abrupt object and camera motion, leading to enhanced localization stability under extreme viewpoint changes.


\paramargin
\subsection{Training} 
\label{sec:loss}

\paragraph{Task loss.}
We define the VQL task loss $L_{\text{task}}(\C,\hat{\C})$ as a function of predicted bounding boxes and the confidence values, $\hat{\C}$, and the ground truth bounding boxes and the occurrence labels $\C$.
Following prior work~\cite{jiang2023vqloc}, the task loss consists of a box regression loss and an object score loss. 
The box regression loss combines $L_1$ loss and the generalized intersection over union (GIoU) loss~\cite{rezatofighi2019generalized}.
For the object score loss, we employ focal loss~\cite{lin2017focal}.

\paramargin
\paragraph{Consistency training loss.}
\label{sec:ctloss}
To enhance temporal stability, we introduce Consistency Training (CT) loss as illustrated in \figref{overview} (c).
Given the predictions $\hat{\C}$ from the original video $\Ivideo$ and the predictions $\hat{\C}'$ from the frame-reordered video $\mathbf{I}_\text{MotionAug}$ generated by MotionAug, CT loss penalizes discrepancies between them: $L_{\text{CT}} = L_{\text{task}}(\hat{\C},\hat{\C'})$.
By enforcing CT loss, the model learns to consistently predict across different motion speeds, enhancing robustness to abrupt viewpoint and object movement variations.

\paramargin
\paragraph{Total loss.}
We define the total loss function for training as follows:
\begin{equation} 
\label{eq:total_loss_eq}
    L = \beta L_{\text{task}}(\C,\hat{\C}) + \gamma L_{\text{task}}(\C',\hat{\C}') + \lambda L_{\text{CT}} + \mu L_{\text{TAG}},
\end{equation} 
where $\beta$, $\gamma$, $\lambda$, and $\mu$ are hyperparameters that control the relative contributions of each loss term. 
We define $L_{\text{TAG}}$ as an entropy-based loss function, which helps refine the attention mechanism.
For further details, please refer to the supplementary material.

\paramargin

%% file: figure/fig_overview.tex
\begin{figure}[t]
\includegraphics[width=\linewidth]{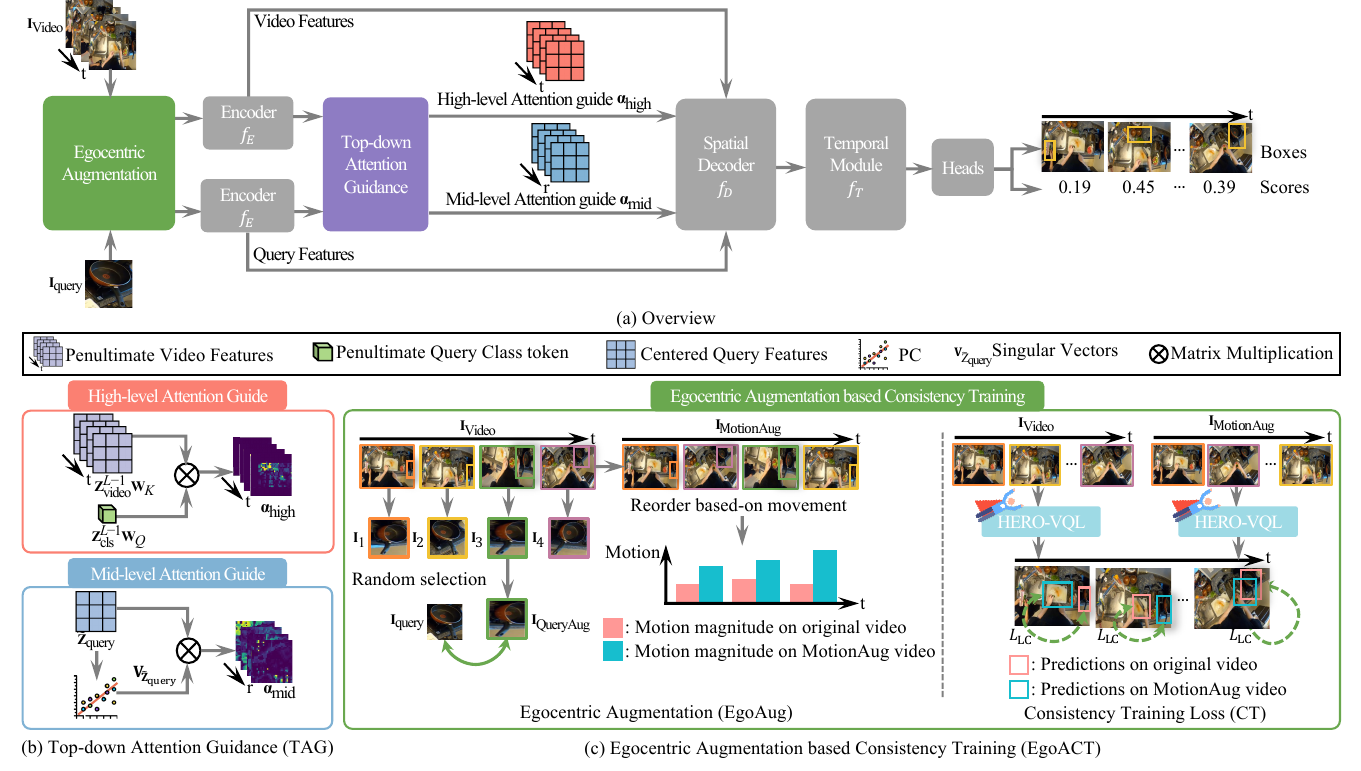}
\vspace{-2em}
\figcaption{Overview of \ours{}}
{
    (a) Given a video and a query image, we extract feature vectors using a pre-trained visual encoder. 
    We feed the feature vectors through a spatial decoder, followed by a temporal module and a prediction head outputs per-frame bounding boxes and scores.
    (b) TAG guides the spatial decoder’s attention using a high-level cue to capture the overall query context and a mid-level cue to enhance the understanding of fine-grained object parts.
    (c) To enable robust matching, we replace the query image with a randomly selected corresponding object instance from the ground-truth. 
    We reorder video frames based on object movement magnitude to simulate abrupt viewpoint changes. 
    We also enforce temporal consistency, improving localization stability in egocentric videos.
}
\vspace{-1em}
\label{fig:overview}
\end{figure}

%% file: 5_result.tex
\section{Experimental Results}
\label{sec:results}
In this section, we conduct experiments to address the following research questions:
(1) Does \ours{} effectively stabilize object localization across challenging egocentric visual query localization scenarios? (\secref{sota})
(2) Are the hierarchical attention and egocentric-specific training strategies effective in egocentric VQL? (\secref{ablation_study})
(3) How much each of the key components, \ie EgoACT and TAG, contributes to performance gain? (\secref{ablation_study}) 
(4) What information does TAG capture to support robust localization? (\secref{qualitative})
To answer these questions, we first provide dataset details in \secref{results_datasets}, the evaluation metrics in \secref{results_metrics}.
We put implementation details in the supplementary material.

\vspace{-0.5em}
\subsection{Datasets}
\label{sec:results_datasets}
\vspace{-0.5em}
We evaluate \ours{} on Visual Queries 2D Localization (VQ2D)~\cite{grauman2022ego4d} dataset.
VQ2D, part of the Ego4D project, comprises egocentric videos with frequent camera motion and occlusions, with an average video length of 130 seconds.
This dataset provides an extensive evaluation setup for \ours{}, covering egocentric object localization.

\paramargin
\subsection{Evaluation Metrics}
\label{sec:results_metrics}
\vspace{-0.5em}
To evaluate VQ2D, we follow the Ego4D~\cite{grauman2022ego4d} and use the following metrics: \tap{}, \stap{}, Rec. \%, and Succ.
Specifically, \tap{} and \stap{} denote the average precision values thresholded by spatial and spatio-temporal intersection over union (IoU), respectively. 
The recovery (Rec. \%) is frame-level recall with a spatial IoU threshold value of 0.5, while the success rate (Succ.) represents spatio-temporal precision with a low IoU threshold value of 0.05. 

\paramargin
\subsection{Comparison with the State-of-the-Art on VQ2D}
\label{sec:sota}
\vspace{-0.5em}

In this section, we evaluate the performance of \ours{} and state-of-the-art methods, as shown in \tabref{main}.
\ours{} achieves state-of-the-art results with the highest \tap{} of 0.38 and \stap{} of 0.28, surpassing the second-best method, VQLoC, by 7 points and 6 points, respectively.
These results demonstrate the strong performance of the proposed method in challenging egocentric visual query localization scenarios.
Moreover, \ours{} achieves state-of-the-art performance on the test set with a significant margin, further confirming its robustness and generalization capability.
By leveraging the proposed top-down attention guidance and motion-aware augmentations tailored for egocentric scenarios, \ours{} effectively stabilizes object localization, even under abrupt camera movements and occlusions.

\input{table/main_vq2d}

\paramargin
\subsection{Ablation Study}
\label{sec:ablation_study}
\vspace{-0.5em}

We conduct ablation studies to validate the effectiveness of key components and examine design choices in \ours{}.
To ensure a fair comparison, we train and evaluate all models under identical configurations, modifying only the specific component or design choice being tested in each experiment.
We report \tap{}, \stap{}, recovery, and success rate as percentages on the VQ2D validation set to provide a comprehensive performance evaluation.

\paramargin
\paragraph{Effect of TAG and EgoACT.}
To evaluate the two main components, we conduct an ablation study on two key aspects of \ours{}:
(i) top-down attention guidance (TAG), motivated by the hierarchical nature of human visual processing~\cite{Reverse_hierarchy, view_from_top}, and
(ii) egocentric augmentation based consistency training (EgoACT), which enhances robustness to motion and appearance variation.
As shown in \tabref{ablation} (a), ablating EgoACT results in a notable performance drop: \tap{} by 2.8 points and \stap{} by 1.1 points.
The results highlight the importance of motion-aware augmentations and consistency-enforcing strategies for handling appearance variations, abrupt motion, and partial occlusions.
Ablating TAG results in 4.1 points drop in \tap{} and 1.8 points drop in \stap{}, demonstrating its effectiveness in refining localization through structured top-down attention.
Finally, ablating both TAG and EgoACT leads to a significant performance degradation across all metrics (3.1–8.3 points).
The results confirm that the two components work synergistically to enhance visual query localization in challenging egocentric scenarios.

\input{table/ablation}

\paramargin
\paragraph{Effect of high-level and mid-level attention guide.}

We evaluate the efficacy of high-level and mid-level attention guide of TAG in \tabref{ablation} (b). 
The results indicate that both guides significantly enhance performance:
High-level attention guide improves \tap{} by 1.9 points and \stap{} by 1.5 points, demonstrating its role in capturing global object context.
Mid-level attention guide contributes a 3.5 points gain in \tap{} and a 2.2 points gain in \stap{}, highlighting its importance in fine-grained object localization.
When both components are combined, they produce a synergistic effect, achieving a strong overall performance of 37.5\% in \tap{} and 27.6\% in \stap{}, validating the effectiveness of top-down attention guidance.

\paramargin
\paragraph{Effect of query selection strategies in QueryAug.} 
To investigate different design choices for QueryAug, we conduct an ablation study and show the results in \tabref{ablation} (c). 
We compare three query replacement strategies and two traditional augmentations: (1) random replacement, (2) most similar instance, (3) least similar instance, (4) photometric augmentation, and (5) geometric augmentation.
QueryAug outperforms traditional augmentations, indicating that leveraging diverse object instances is more effective than low-level pixel perturbations in egocentric settings.
Within QueryAug, the random replacement achieves the best performance by encouraging learning from a broader range of query–GT pairs.

\paramargin
\paragraph{Effect of frame reordering strategies in MotionAug.} 
To investigate different design choices in MotionAug, we conduct an ablation study and show the results in \tabref{ablation} (d).
We compare three frame reordering strategies: (1) maximizing object displacement, (2) random reordering, and (3) no reordering (w/o MotionAug).
Among these, maximizing object displacement achieves the best performance, demonstrating its effectiveness in enhancing localization robustness under extreme viewpoint changes and abrupt motions.

\paramargin
\paragraph{Effect of consistency training loss.} 
As shown in \tabref{ablation} (e), incorporating CT loss improves \tap{} by 1.5 and \stap{} by 1.0 points, confirming its effectiveness in enhancing temporal stability.
By enforcing consistency between predictions on the original and reordered video sequences, \ours{} becomes more robust to appearance variations, abrupt motions, and partial occlusions, improving localization stability in challenging egocentric scenarios.

\paramargin
\paragraph{Effect of backbone architectures.}
We evaluate whether \ours{} performs consistently across different backbone architectures in \tabref{ablation} (f).
We conduct experiments with DINOv2~\cite{oquab2024dinov} ViT-B/14 and CLIP~\cite{radford2021learningtransferablevisualmodels} ViT-L/14.
\ours{} consistently outperforms the baseline, demonstrating its effectiveness regardless of the backbone architecture.

\paramargin
\subsection{Qualitative Evaluation}
\input{figure/fig_qual_tag_main}
\label{sec:qualitative}
In \figref{tag_main}, We visualize TAG’s high-level and mid-level attention guides to understand what types of information they capture.
The high-level attention guide attends more strongly to patch tokens in the video features that are associated with the object.
While the attention is broadly distributed across candidate regions, the patch token corresponding to the query object receives particularly high attention.
Additionally, the mid-level attention guide attends to specific object parts in the query feature.
The difference in visualization reflects their roles in the decoder.
The high-level attention guide is added to the self-attention map with video-only inputs, whereas the mid-level attention guide is applied to the cross-attention map, where video features act as queries and the query image features serve as keys and values.
These results highlight the effectiveness of TAG in enabling robust visual query localization, effectively addressing varied challenges in egocentric videos.

%% file: table/main_vq2d.tex
\begin{table}[t]
\centering
\resizebox{0.7\columnwidth}{!}{
\begin{tabular}{l c cccc c cccc}
\toprule
& \multicolumn{4}{c}{Validation Set} & \multicolumn{4}{c}{Test Set} 
\\
\cmidrule(lr){2-5}
\cmidrule(lr){6-9}
Method & $\text{tAP}_{25}$ & $\textbf{stAP}_{25}$ & Rec. \% & Succ.
& $\text{tAP}_{25}$ & $\textbf{stAP}_{25}$  & Rec. \% & Succ.
\\
\midrule
SiamRCNN~\cite{grauman2022ego4d}
& 0.20 & 0.12 & 32.2 & 39.8
& 0.21 & 0.13 & 34.0 & 41.6
\\
NFM~\cite{xu2022negative}
& 0.26 & 0.19 & 37.9 & 47.9
& 0.24 & 0.17 & 38.6 & 47.7
\\
CocoFormer~\cite{xu2023my}
& 0.26 & 0.19 & 37.7 & 47.7
& 0.26 & 0.18 & 43.2 & 48.1
\\
VQLoC ~\cite{jiang2023vqloc}
& 0.31 & 0.22 & \textbf{47.1} & 55.9
& 0.32 & 0.24 & 45.1 & 55.9
\\
\ours{} (ours) 
& \textbf{0.38} & \textbf{0.28} & 44.9 & \textbf{61.1}
& \textbf{0.37} & \textbf{0.28} & \textbf{45.3} & \textbf{60.7}
\\
\bottomrule
\end{tabular}
}
\caption{
    \tb{Comparison with state-of-the-art on VQ2D~\cite{grauman2022ego4d}.} 
    We report the \tap{}, \stap{}, recovery (Rec. \%), and the success rate (Succ.) on the validation set and the test set. 
}
\label{tab:main}
\vspace{-1em}

\end{table}

%% file: table/ablation.tex
\begin{table*}[t]
\centering

\mpage{0.47}{
\resizebox{0.78\columnwidth}{!}{
\begin{tabular}{l c cccc}
\toprule
Components && $\text{tAP}_{25}$ & $\textbf{stAP}_{25}$ & Rec. \% & Succ.
\\
\midrule
\ours{} && 
\textbf{37.5} & \textbf{27.6} & 44.9 & \textbf{61.1}
\\
w/o. EgoACT && 
34.7 & 26.5 & \textbf{45.1} & 59.8
\\
w/o. TAG && 
33.4 & 25.8 & 43.6 & 59.1
\\
Baseline && 
29.2 & 21.9 & 41.8 & 54.8
\\
\bottomrule
\end{tabular}
}
}
\hfill
\mpage{0.51}{
\resizebox{0.8\columnwidth}{!}{
\begin{tabular}{c c cccc}
\toprule
High-level & Mid-level  & $\text{tAP}_{25}$ & $\textbf{stAP}_{25}$ & Rec. \% & Succ.
\\
\midrule
\cmark & \cmark & 
\textbf{37.5} & \textbf{27.6} & 44.9 & \textbf{61.1}
\\
\cmark & \graycross & 
34.4 & 25.7	& 43.6 & 59.9
\\
\graycross & \cmark &
35.6 & 26.1 & \textbf{45.1} & 59.6
\\
\bottomrule
\end{tabular}
}
}
\\
\mpage{0.49}{\fontsize{8}{10}\selectfont (a) Effect of TAG and EgoACT components.}\hfill
\mpage{0.49}{\fontsize{8}{10}\selectfont (b) Effect of high-level and mid-level guide in TAG.}\hfill
\\
\mpage{0.48}{
\resizebox{0.9\columnwidth}{!}{
\begin{tabular}{l c cccc}
\toprule
Query Aug. Method && $\text{tAP}_{25}$ & $\textbf{stAP}_{25}$ & Rec. \% & Succ.
\\
\midrule
Random replacement && 
\textbf{37.5} & \textbf{27.6} & \textbf{44.9} & \textbf{61.1}
\\
Least similar instance && 
34.4	&25.5	&45.6	&59.6
\\
Most similar instance && 
33.9	&25.2	&43.9	&59.1
\\
\midrule
Photometric Aug. && 
30.2	& 22.0	&40.2	&56.2
\\
Geometric Aug. && 
29.8	&22.2	&39.3	&56.6
\\
\bottomrule
\end{tabular}
}
}
\hfill
\mpage{0.49}{
\resizebox{\columnwidth}{!}{
\begin{tabular}{l  cccc}
\toprule
Strategy & $\text{tAP}_{25}$ & $\textbf{stAP}_{25}$ & Rec. \% & Succ.
\\
\midrule
MotionAug &
\textbf{37.5} & \textbf{27.6} & 44.9 & \textbf{61.1}
\\
MotionAug Random &
34.9 & 26.2 & 44.8 & 59.7
\\
w/o. MotionAug &
34.2 & 26.3 & \textbf{45.4} & 59.6 
\\
\bottomrule
\end{tabular}
}
}
\\
\mpage{0.51}{\fontsize{8}{10}\selectfont (c) Effect of query selection strategies.}\hfill
\mpage{0.47}{\fontsize{8}{10}\selectfont (d) Effect of frame reordering strategies.}\hfill
\\
\mpage{0.47}{
\resizebox{0.85\columnwidth}{!}{
\begin{tabular}{c  cccc}
\toprule
CT loss & $\text{tAP}_{25}$ & $\textbf{stAP}_{25}$ & Rec. \% & Succ.
\\
\midrule
\cmark & 
\textbf{37.5} & \textbf{27.6} & 44.9 & \textbf{61.1}
\\
\graycross  &
36.0 & 26.6 & \textbf{45.6} & 60.5
\\
\bottomrule
\end{tabular}
}
}
\hfill
\mpage{0.48}{
\resizebox{0.9\columnwidth}{!}{
\begin{tabular}{ll cccc}
\toprule
Method & Backbone  & $\text{tAP}_{25}$ & $\textbf{stAP}_{25}$ & Rec \% & Succ.
\\
\midrule
Baseline & \multirow{2}{*}{DINOv2~\cite{oquab2024dinov}} 
& 29.2 & 21.9 & 41.8 & 54.8
\\
\ours{} & & \textbf{37.5} & \textbf{27.6} & \textbf{44.9} & \textbf{61.1}
\\
\midrule
Baseline & \multirow{2}{*}{CLIP~\cite{radford2021learningtransferablevisualmodels}} 
& 25.0 & 16.5 & 34.7 & 50.5
\\
\ours{} &
& 31.9 & 20.5 & 36.5 & 56.5
\\
\bottomrule
\end{tabular}
}
}
\\
\mpage{0.48}{\fontsize{8}{10}\selectfont (e) Effect of CT loss.}\hfill
\mpage{0.48}{\fontsize{8}{10}\selectfont (f) Effect of backbone architectures.}\hfill
\\
\vspace{-0.5em}
\caption{\textbf{Ablation study.} 
To validate the effect of each component, we show the results on the VQ2D validation set. We report \tap{}, \stap{}, recovery, and success rate as percentages.  
}

\vspace{-1.5em}

\label{tab:ablation}
\end{table*}

%% file: figure/fig_qual_tag_main.tex
\begin{figure}[t]
    \includegraphics[width=\textwidth]{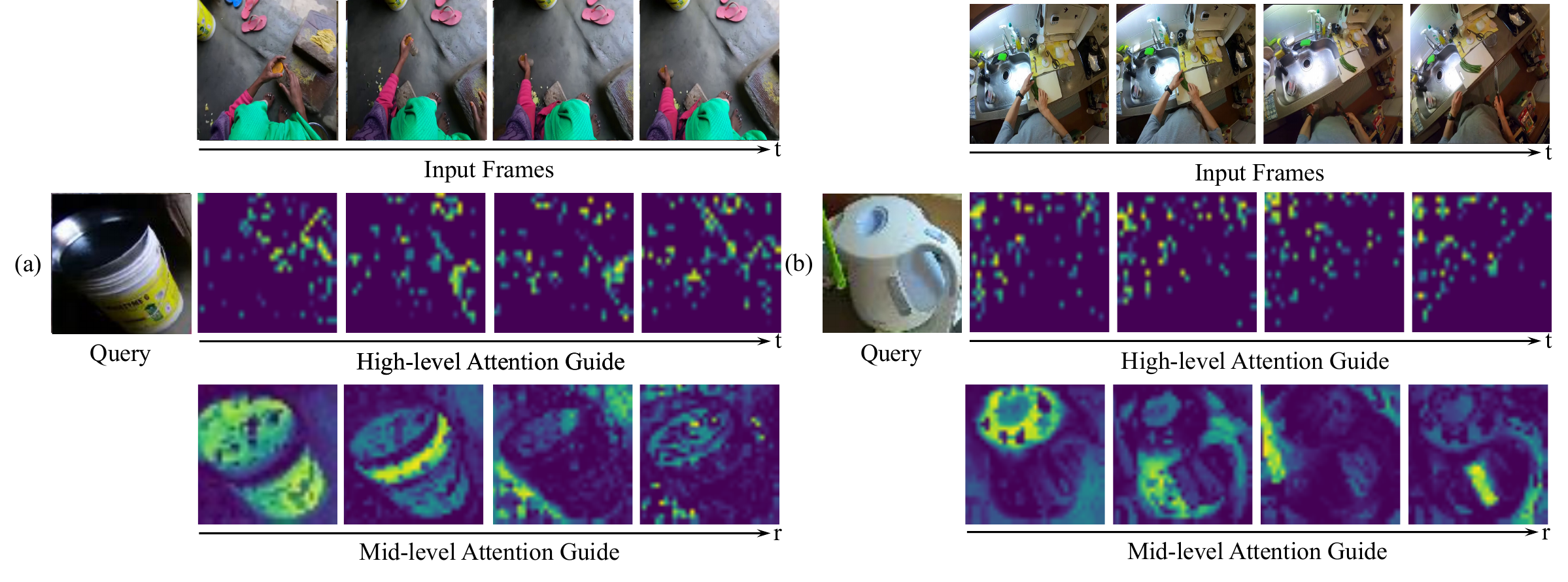}
    \vspace{-1.5em}
    \figcaption{Visualization of TAG}{
    We show the query image and four frames from each video with TAG's high-level and mid-level attention guides.
    The high-level attention guide emphasize the region of the target object and the mid-level attention guide attends to specific object parts in the query feature.
    They improve the model's ability to localize objects accurately, under varying perspectives or partial visibility.
    }
    \label{fig:tag_main}
    \vspace{-1em}
\end{figure}

%% file: 6_conclusion.tex
\section{Conclusions}
\label{sec:conclusions}
\vspace{-0.5em}
In this paper, we address challenges of egocentric visual query localization (VQL) task, which include appearance variations, abrupt motions, and partial occlusions.
We propose \ours{}, a method designed to improve localization robustness through two key components:
(i) TAG refines attention through hierarchical top-down guidance, enhancing object localization under varying perspectives and partial occlusions.
(ii) EgoACT, a training strategy that improves robustness by simulating challenging egocentric scenarios, \ie query replacement and motion-aware frame reordering, and enforcing consistency in localization between the original clip and an augmented clip.
Extensive experiments demonstrate that \ours{} significantly improves robustness in egocentric scenarios and achieves state-of-the-art performance on the challenging VQ2D benchmark.

%% file: 50_supp_content.tex
In this supplementary material, we provide comprehensive dataset/baseline and method details to complement the main paper. We organize the supplementary material as follows:
\begin{enumerate}
    \paramargin
    \item Dataset details
    \paramargin
    \item Implementation details
    \paramargin
    \item Method details
    \paramargin
    \item Comprehensive quantitative results
    \paramargin
    \item Comprehensive qualitative evaluations
\end{enumerate}


\section{Dataset Details}

In this section, we provide a detailed description of Visual Queries 2D localization (VQ2D)~\cite{grauman2022ego4d} dataset.

\paramargin
\paragraph{Details}
VQ2D dataset is a large-scale egocentric video dataset for localization that contains about 6K clips.
The dataset is split into train/val/test sets, with 3.6k/1.2k/1.1k clips and 13.6k/4.5k/4.4k queries, respectively.
Bounding boxes are annotated at the rate of 5 frames per second.
\paramargin
\paragraph{Metrics}
The evaluation of VQ2D is based on four key metrics: \tap{}, \stap{}, Rec \%, and Succ. These metrics are defined as follows:
\begin{enumerate} 
\vspace{-0.5em}
\item \tap{}: Temporal Average Precision (AP) calculated at a threshold of temporal Intersection over Union (tIoU). Specifically, we evaluate \tap{} at a tIoU threshold of 0.25, indicating how well the predicted temporal boundaries align with the ground truth. Higher values suggest better temporal localization.
\paramargin
\item \stap{}: Spatio-temporal Average Precision (AP) calculated at a threshold of spatio-temporal Intersection over Union (stIoU). This metric measures the overlap between the predicted and ground-truth spatio-temporal volumes, capturing both spatial accuracy and temporal alignment.
\paramargin
\item Recovery \% (Rec \%): Frame-level recall, which measures the proportion of frames correctly identified within a video. For a frame to be considered correctly recovered, the spatial IoU between the prediction and ground truth must exceed 0.5.
\paramargin
\item Success rate (Succ.): Spatio-temporal precision, representing the proportion of spatio-temporal predictions that meet a minimum IoU threshold of 0.05. This metric emphasizes identifying whether the prediction captures even a minimal level of overlap with the ground truth.
\end{enumerate}


\vspace{-1em}
\section{Implementation Details}
In this section, we provide details of our experimental setup and implementation on VQ2D dataset~\cite{grauman2022ego4d}.
We conduct all the experiments with 8 RTX 3090s or RTX A5000s.
We implement the proposed method in PyTorch~\cite{paszke2017automatic} and PyTorch Lightning from scratch.

\secmargin
\subsection{Training} 
\vspace{-0.5em}
During training, we divide each untrimmed video into non-overlapping, fixed-length clips of 32 frames at 5 fps.
Each clip contains at least one instance of the target object.
We resize all frames and query images to 448 $\times$ 448 pixels, setting the longer side to 448 and applying zero-padding to the shorter side to maintain the aspect ratio.
We set total loss hyperparameters as: $\beta=0.1\overline{6}$, $\gamma=0.1\overline{6}$, $\lambda=0.\overline{6}$, $\mu=0.1$, and we set the QueryAug probability $p=0.5$. 
We train the model for 106 epochs with a batch size of 24.
We use the AdamW~\cite{loshchilov2017decoupled} optimizer with a learning rate of 0.0003 and a weight decay of 0.005.
We employ a warm-up scheduling for the learning rate for the initial 1,000 iterations.  
We employ DINOv2~\cite{oquab2024dinov} (ViT-B/14) as an encoder and TSM~\cite{lin2019tsm} as a temporal module. 

\secmargin
\subsection{Inference} 
\vspace{-0.5em}
During inference, we split the video in the same way as in training. 
We concatenate the predicted bounding boxes and their corresponding scores across clips to obtain a continuous sequence of predictions throughout the video.
To smooth the score sequence, we apply a median filter with a kernel size of 5.
From the smoothed sequence, we first identify the peak with the highest score, denoted as $\pi_h$.
Then we threshold the score sequence with a value $0.7\pi_h$ to find \emph{candidate intervals} where the target object appears.
For the VQ2D dataset, we consider only the last temporal segment.

\secmargin
\subsection{Box Movement Measurement in MotionAug}
\vspace{-0.5em}
We measure box variation for reordering frames in MotionAug.
Given a set of bounding boxes represented by their coordinates in the form: 
$
\begin{bmatrix}
y_1, x_1, y_2, x_2
\end{bmatrix}.
$
For each bounding box, the following computations are performed:
The center points are calculated as:
\[
c_x = \frac{x_1 + x_2}{2},\quad c_y = \frac{y_1 + y_2}{2}
\]
The width and height are computed as:
\[
w = x_2 - x_1,\quad h = y_2 - y_1
\]
The pairwise differences in center coordinates are:
\[
\Delta x = c_{x_i} - c_{x_j}, \quad \Delta y = c_{y_i} - c_{y_j}
\]
The Euclidean distance between bounding box centers is:
\[
D = \sqrt{\Delta x^2 + \Delta y^2}
\]
The width and height differences between bounding boxes are:
\[
\Delta w = w_i - w_j,\quad  \Delta h = h_i - h_j
\]
To compute scale ratios, we ensure valid width and height values:
\[
V_w = (w_i > 0) \land (w_j > 0), \quad  V_h = (h_i > 0) \land (h_j > 0)
\]
The logarithmic scale ratios are:
\[
S_w = 
\begin{cases} 
\log\left(\frac{w_i}{w_j}\right), & \text{if } V_w \\
0, & \text{otherwise}
\end{cases}
\]
\[
S_h = 
\begin{cases} 
\log\left(\frac{h_i}{h_j}\right), & \text{if } V_h \\
0, & \text{otherwise}
\end{cases}
\]
Finally, the total change $\Delta_{total}$ is computed as:
\[
\Delta_{\text{total}} = D + |\Delta w| + |\Delta h| + |S_w| + |S_h|
\]


\paramargin
\subsection{Hyperparameter Settings}

All experiments and ablation studies use the same settings. For encoders, CLIP~\cite{radford2021learningtransferablevisualmodels} ViT-L/14 processes a batch size of 2 per GPU, while other encoders use a batch size of 3 per GPU.
We summarize the hyperparameters in \tabref{hyperparam}.

\input{table/supp_hyperparam}


\section{Method Details}
\subsection{Baseline Details}
\input{figure/fig_baseline}
We describe the baseline architecture, which excludes all our key components, including TAG and EgoACT, as illustrated in \figref{baseline}.
The architecture consists of a pre-trained encoder $f_E$, a spatial decoder $f_D$, a temporal module and a prediction head.
We extract video features $\Zvideo$ and query feature $\Zquery$ from encoder.
The video features act as the query, while the query feature serves as the key and value, and both are processed by the spatial decoder.
The outputs of the spatial decoder are passed through the temporal module.
Finally, the prediction head predicts a bounding box $\mathbf{B}_i = \{ x_i, y_i, w_i, h_i \}$ and the confidence score $\pi_i$ of $i$-th frame.

\paramargin
\subsection {Ego-motion Augmentation}
To increase camera motion, MotionAug reorders frames, transforming video into a more dynamic scenario. We select the first ground-truth frame and measure box movement from the other frames. Then, we choose the frame with the maximum displacement and place it next to the first frame. We repeat this process for each frame to amplify viewpoint changes. 
We summarize MotionAug in \algref{alg:emaug}.
\begin{algorithm}[t]
\caption{Ego-motion Augmentation (MotionAug)}
\label{alg:emaug}
\SetKwInput{Input}{Input}
\SetKwInput{Output}{Output}

\Input{RGB video $\Ivideo = \{I_1, I_2, ..., I_T\}$, \\
GT frames $\mathbb{F}_\text{GT} \subseteq \Ivideo$}
\Output{Reordered video $\Ivideo'$}

Initialize $\Ivideo' = \Ivideo$ \tcp*[f]{Copy video}  \\
Initialize $\mathbb{F}_\text{GT}' = [\ ]$ \tcp*[f]{Reordered GT frames}

Select first GT frame $F \in \mathbb{F}_\text{GT}$ and append to $\mathbb{F}_\text{GT}'$
Remove $F_1$ from $\mathbb{F}_\text{GT}$

\While{$\mathbb{F}_\text{GT}$ is not empty}{
    Select $F^*$ with max displacement from last chosen frame:
    \begin{equation}
        F^* = \argmax_{F' \in \mathbb{F}_\text{GT}} \text{Displacement}(F_{\text{last}}, F')
    \end{equation}
    Append $F^*$ to $\mathbb{F}_\text{GT}'$
    Remove $F^*$ from $\mathbb{F}_\text{GT}$
}

Replace GT frames in $\Ivideo'$ with $\mathbb{F}_\text{GT}'$

\Return $\Ivideo'$
\end{algorithm}

\subsection{High-level Attention Guide}
\input{figure/fig_detail_GAG}
As shown in \figref{detail_gag}, we illustrate the sources of $\Zvideo^{L-1}\W_K$ and $\Zcls^{L-1}\W_Q$ for computing the high-level attention guide $\ahigh$.
We extract $\Zvideo^{L-1} \in \mathbb{R}^{M \times D}$ and $\Zcls^{L-1} \in \mathbb{R}^{D}$ from the $(L-1)$th layer, referred to as the penultimate layer of the encoder, where $D$ is the token dimension.
$\Zvideo^{L-1}$ represents the $M$ video patch tokens, and $\Zcls^{L-1}$ corresponds to the class token of the query image.

To compute high-level attention guide $\ahigh$, we use the query projection matrix $\W_Q \in \R^{D \times D}$ and the key projection matrix $\W_K \in \R^{D \times D}$ from the self-attention mechanism of the same layer.
Before applying the projection, we refine $\Zvideo^{L-1}$ to repair attention scores by addressing high-norm tokens that often dominate attention.
Specifically, we replace these high-norm tokens with repaired vectors, computed as the mean of their neighboring tokens.

After repairing $\Zvideo^{L-1}$, we compute high-level attention guide $\ahigh$ by multiplying the key vector $\Zvideo^{L-1}\W_K \in \R^{M \times D}$ with the query vector $\Zcls^{L-1}\W_Q \in \R^{D}$. We then center the values prior to applying the sigmoid function $\sigma(\cdot)$ to normalize the attention scores.

\paramargin
\paragraph{Attention score repair.}
We observe that a few high-norm vectors tend to cluster in the very top-left region of the attention maps.
These high-norm vectors often dominate other tokens, leading to non-informative guidance.
We simply replace each high-norm token $\mathbf{z}_\text{hn}$ with a repaired vector $\mathbf{z}_\text{repair}$, whose magnitude and direction are set to the means of the neighboring tokens of $\mathbf{z}_\text{hn}$.
By this simple fix, we prevent the attention score distribution from becoming excessively concentrated in the non-informative region.

\input{figure/fig_qual_tag}
\paramargin
\subsection{Visualize Top-down Attention Guidance}
We visualize TAG’s high-level attention guide $\alpha_\text{high}$ and mid-level attention guide $\alpha_\text{mid}$ to understand what types of information they capture.
The two guides play complementary roles: 
the high-level guide $\alpha_\text{high}$ provides global semantic cues from the query’s class token, which are injected into self-attention to bias the video patches toward object-relevant regions. 
In contrast, the mid-level guide $\alpha_\text{mid}$ captures fine-grained object parts through principal component maps and is integrated into cross-attention, where each head aligns a distinct part of the query with video patches. 
Together, they enable a progressive refinement from global context to local details, leading to more robust localization under viewpoint changes and occlusions.


\paramargin
\subsection{Top-down Attention Guidance in the Spatial Decoder}
We describe how the spatial decoder incorporates the TAG into its self-attention and cross-attention mechanisms.

\paramargin
\paragraph{Self-Attention with high-level attention guide}
We first incorporates the high-level attention guide $\ahigh$ into the self-attention mechanism as a guidance. 
Specifically, this attention mask modifies the self-attention computation as follows:
\begin{equation}
\label{eq:docoder_w_hag_eq}
    \mathbf{A}_\text{self} = \texttt{Softmax} \left ( \frac{\Q_{SA}\K^\top_{SA}}{\sqrt{D}} + \ahigh \right ),
\end{equation}
\noindent
where $\Q_{SA}$ and $\K_{SA}$ are the $M \times D$ query and key projection matrices of the decoder self-attention layer.
Guided by \( \ahigh \), the spatial decoder is able to focus on video regions more relevant to the query object.

\paramargin
\paragraph{Cross-Attention with mid-level attention guide.}
Before applying the mid-level attention guide, we replicate the $r$-th column vector of $\S$, $\s_r\in \R^N$, $M$ (the number of video patch tokens) times:  
\vspace{-0.1em}
\begin{equation}
    \alpha_\text{mid}^{r'} = \begin{bmatrix}
    \alpha_\text{mid}^r, \alpha_\text{mid}^r, \cdots , \alpha_\text{mid}^r
    \end{bmatrix} \in [0,1)^{M \times N}.
\end{equation}
Then the spatial decoder incorporates the mid-level attention guide $\alpha_\text{mid}^r$ into the multi-head cross-attention mechanism, where the number of principal components matches the number of attention heads $N_\text{head}=R$, allowing each component to guide a separate attention head:  
\begin{equation}
    \mathbf{A}^{r}_\text{cross} = \texttt{Softmax} \left( \frac{\Q^r_{CA}\K^{r \top}_{CA}}{\sqrt{D/N_\text{head}}} + \alpha_\text{mid}^{r'} \right),
\end{equation}
where $\Q_{CA}^r$ is the $M \times D/N_\text{head}$ query and $\K_{CA}^r$ is the $N \times D/N_\text{head}$ key matrices of the $r$-th head in the cross-attention layer.
Guided by $\alpha_\text{mid}^{r'}$, each head focuses on different object parts, refining object localization under partial visibility issues and viewpoint changes.


\paramargin
\subsection{Top-down Attention Guide Loss}
\label{sec:tagloss}
We introduce Top-down Attention Guide (TAG) loss to ensure structured attention across object parts, encouraging the decoder to learn diverse object representations.
We construct TAG score maps by projecting the centered query feature $\tZquery$ onto the principal components of the object-aware features $\Y$:

\begin{equation} 
\label{eq:tag_score_eq}
    \S_{\text{TAG}} = \{ \mathbf{s}_\text{TAG}^i \in [0, 1)^{R} \}_{i=1}^{N} = \phi \left( \tZquery \V_{\Y } \right) .
\end{equation}
To ensure distinct object representations, we define TAG loss as a combination of token-wise entropy loss $\ltoken$ and negative map-wise entropy loss $\lmap$.  
\begin{align} 
\label{eq:token_loss_eq}
    \ltoken &= \frac{1}{N}\sum_{i=1}^N \text{H}\left( \texttt{Softmax} \left ( \mathbf{s}_{\text{TAG}}^{i} \right ) \right), \\
\label{eq:map_loss_eq}
    \lmap &= -\text{H} \left ( \texttt{Softmax} \left ( \frac{1}{N} \sum_{i=1}^N \mathbf{s}_{\text{TAG}}^{i} \right ) \right ).
\end{align}
Here, $\text{H}(\cdot)$ denotes the entropy function.
The token-wise entropy loss $\ltoken$ minimizes the entropy of each token’s score across different maps, encouraging diverse feature activations at each spatial location.  
Meanwhile, the negative map-wise entropy loss $\lmap$ maximizes the entropy among globally average-pooled maps, promoting diverse object-level representations by balancing global attention responses and preventing over-reliance on dominant components.

The final TAG Loss is defined as:
\begin{equation} 
\label{eq:tag_loss_eq}
    L_{\text{TAG}} = \lambda_{\text{token}}\ltoken + \lambda_{\text{map}}\lmap,
\end{equation} 
where $\lambda_{\text{token}}$ and $\lambda_{\text{map}}$ are hyperparameters balancing the two loss terms.  
By encouraging a structured transition from global semantics to fine-grained details, TAG Loss improves object localization robustness against occlusions and viewpoint variations.


\section{Comprehensive Quantitative Results}
In this section, we provide additional quantitative results on the VQ2D dataset~\cite{grauman2022ego4d} to complement the main paper. 
We demonstrate 
(1) the effect of attention score repair in \secref{supp_repair}.
(2) QueryAug hyperparameter analysis in \secref{supp_vuaug_p}.

All models are trained under identical configurations, varying only the specific component or design choice being tested.
We report \tap{}, \stap{}, recovery (Rec. \%), and success rate (Succ.) as percentages.

\paramargin
\subsection{Effect of attention score repair}
\label{sec:supp_repair}
\input{table/ablation_repair}
We investigate the effect of repairing attention scores in high-level attention guide in \tabref{abl_gag_repair}.
Applying the attention score repair improves 3.2 points in \tap{} and 2.4 points in \stap{}.
By replacing high-norm tokens with the mean of their neighbors, we prevent the attention score distribution from becoming excessively concentrated in the non-informative region and resulting in performance improvement.

\paramargin
\subsection{QueryAug hyperparameter analysis}
\label{sec:supp_vuaug_p}
\input{table/ablation_vqaug_p}
In \tabref{abl_vqaug_p}, we study the effect of the probability for QueryAug.
We achieve the best performance when the probability is set to 0.5.
This suggests that a low probability limits the model’s exposure to diverse query-GT pairs, leading to degraded performance.


\section{Comprehensive Qualitative Evaluation}
To better understand the effectivness of {\ours}, we present qualitative examples in three challenging scenarios. 
In \figref{partial}, \ours{} successfully localizes the object even when it is partially visible, while VQLoC struggles. 
In \figref{appearance}, while VQLoC~\cite{jiang2023vqloc} fails to localize object instances, {\ours} accurately localizes the object despite rapid camera movement.
In \figref{lemonjin}, without additional fine-tuning, \ours{} effectively localizes the query object in an unseen video with conditions significantly different from the training data.
In contrast, VQLoC produces imprecise predictions.
These results highlight the effectiveness of TAG and EgoACT in enabling robust and generalizable visual query localization, successfully addressing various challenges in egocentric videos.
\input{figure/fig_qual_partial}
\input{figure/fig_qual_apperance}
\input{figure/fig_qual_lemonjin}

%% file: table/supp_hyperparam.tex
\begin{table}[t]
\centering
\resizebox{0.5\columnwidth}{!}{
\begin{tabular}{lc cc}
\toprule
Config  & VQ2D~\cite{grauman2022ego4d}
\\
\midrule
Optimizer & AdamW~\cite{loshchilov2017decoupled}
\\
Backbone matmul precision & TensorFloat32
\\
Learning rate & 3e-3
\\
Weight decay & 0.005
\\
{Optimizer momentum~\cite{chen2020generative}} & {$\beta_1, \beta_2{=}0.9, 0.999$}
\\
Per-GPU batch size & 3
\\
Update frequency & 1
\\
Learning rate schedule & {Linear}
\\
Warmup iterations & 100
\\
Training epochs &  106
\\
Flip augmentation & \cmark 
\\
RandomResizedCrop & \cmark
\\
Brightness jitter & 0.4 
\\
Contrast jitter & 0.4
\\
Saturation jitter & 0.3 
\\
$\beta$ & $0.1\overline{6}$ 
\\
$\gamma$ & $0.1\overline{6}$
\\
$\lambda$ & $0.\overline{6}$
\\
$\mu$ & 0.1
\\
QueryAug probability $p$ & 0.5
\\

\bottomrule
\end{tabular}
}
\caption{
    \tb{Hyperparameters used for training on VQ2D dataset.} 
}
\label{tab:hyperparam}
\tabcapmargin
\end{table}

%% file: figure/fig_baseline.tex
\begin{figure}[t]
    \centering
    \includegraphics[width=\linewidth]{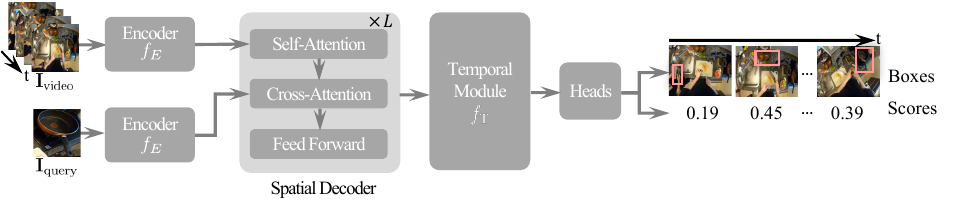}
    \figcaption{Architecture visualization of baseline}
    \figcapmargin
    \label{fig:baseline}
\end{figure}

%% file: figure/fig_detail_GAG.tex
\begin{figure}[t]
    \centering
    \includegraphics[width=0.8\linewidth]{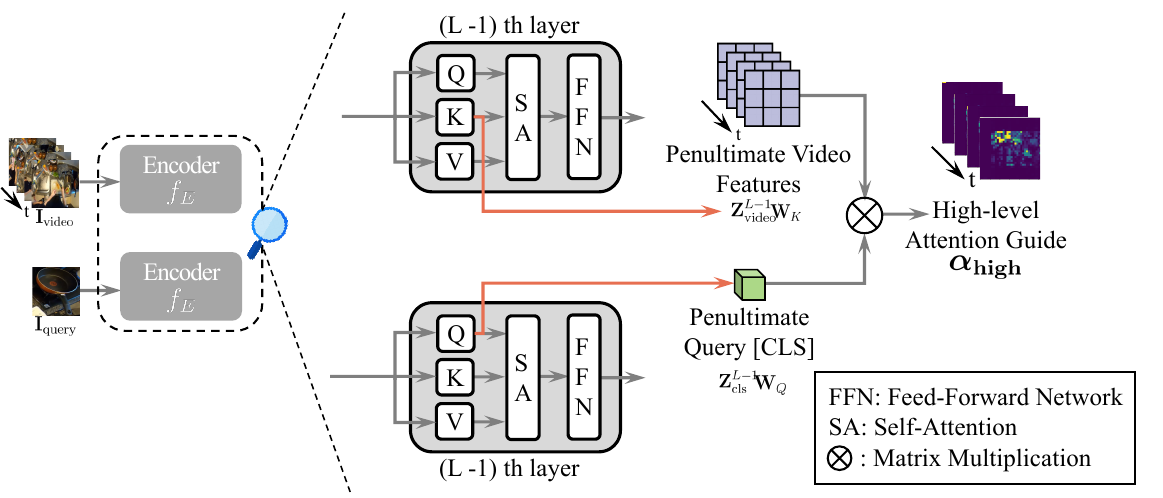}
    \figcaption{Visualization of additional details in high-level attention guidance}
    \figcapmargin
    \label{fig:detail_gag}
    \vspace{-1em}
\end{figure}

%% file: figure/fig_qual_tag.tex
\begin{figure*}[t]
    \centering
    \includegraphics[width=\linewidth]{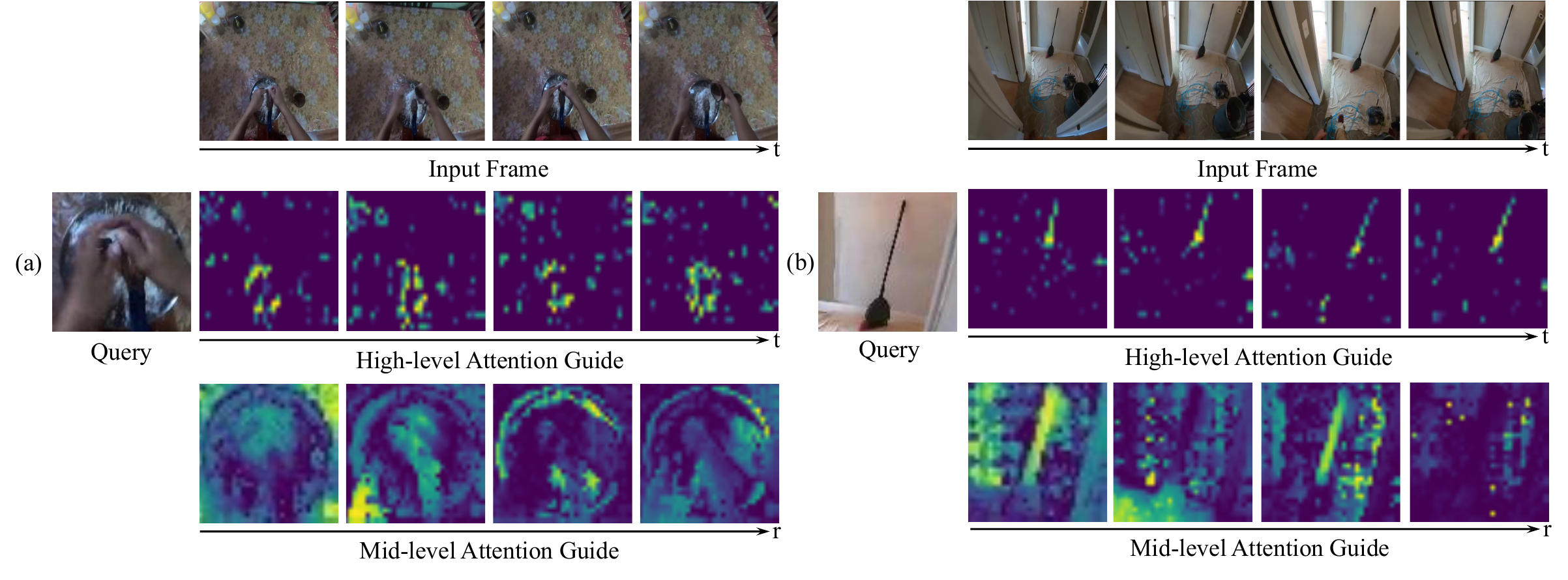}

    \vspace{-0.5em}
    \figcaption{\tb{Visualization of TAG}}
    \figmargin
    \label{fig:tag}
\end{figure*}

%% file: table/ablation_repair.tex
\begin{table}[h]
\centering
\resizebox{0.6\columnwidth}{!}{
\begin{tabular}{l c cccc}
\toprule
Method && $\text{tAP}_{25}$ & $\textbf{stAP}_{25}$ & Rec. \% & Succ.
\\
\midrule
\ours{} with repair && 
\textbf{37.5} & \textbf{27.6} & \textbf{44.9} & \textbf{61.1}
\\
\ours{} without repair && 
34.3 & 25.2 & 44.3 & 59.5
\\
\bottomrule
\end{tabular}
}
\caption{
    \tb{Effect of Attention Score Repair.} 
}
\label{tab:abl_gag_repair}
\end{table}

%% file: table/ablation_vqaug_p.tex
\begin{table}[ht]
\centering
\resizebox{0.6\columnwidth}{!}{
\begin{tabular}{l c cccc}
\toprule
Probability p && $\text{tAP}_{25}$ & $\textbf{stAP}_{25}$ & Rec. \% & Succ.
\\
\midrule
1.0 && 
34.1	&24.4	&41.5   &57.9
\\
0.75 && 
33.6	&25.4	&44.6	&59.1
\\
0.5 &&
\textbf{37.5} & \textbf{27.6} & \textbf{44.9} & \textbf{61.1}
\\
0.25 && 
33.9	&25.7	&43.7	&59.2
\\
w/o. QueryAug && 
33.9	&25.5	&44.1	&58.4
\\
\bottomrule
\end{tabular}
}
\caption{
    \tb{QueryAug hyperparameter analysis.} 
}
\label{tab:abl_vqaug_p}
\tabcapmargin
\end{table}

%% file: figure/fig_qual_partial.tex
\definecolor{qgreen}{HTML}{41A24D}
\definecolor{qred}{HTML}{EB321A}
\definecolor{qblue}{HTML}{0025F6}

\begin{figure*}[p]
    \centering
    \includegraphics[width=\linewidth]{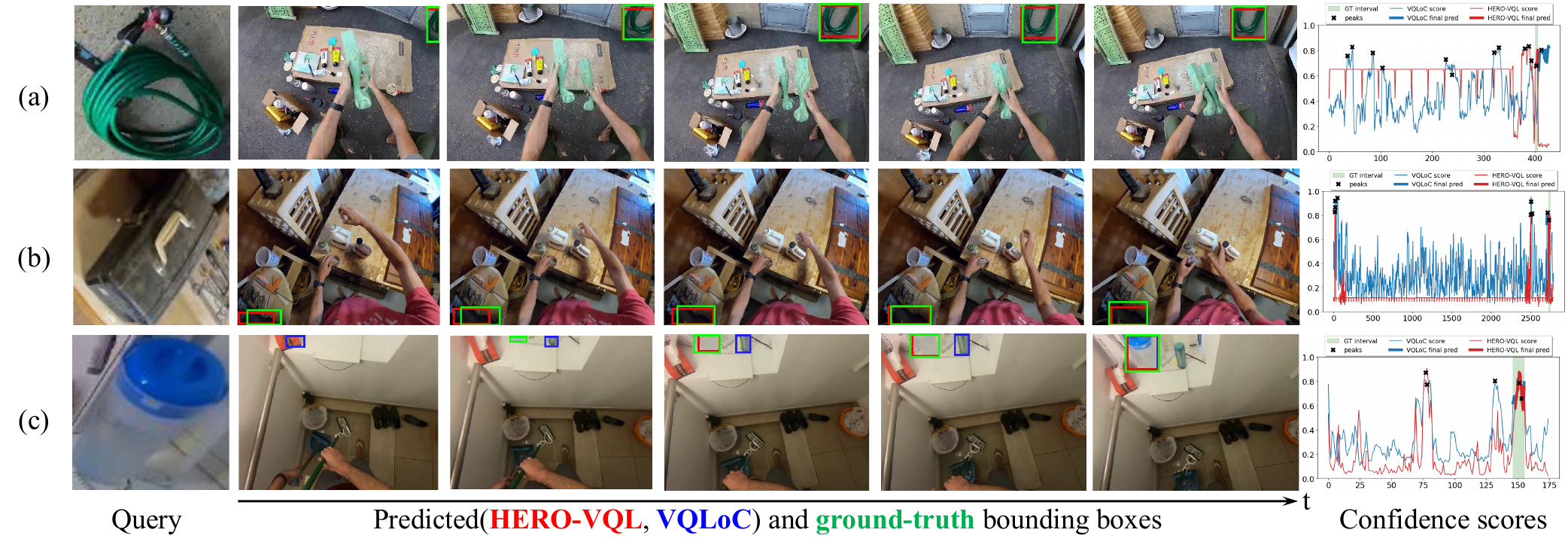}
    \vspace{-1em}
    \figcaption{Qualitative examples in videos with partial object visibility}{
    In each row, we show the query image and five frames from each video with predicted bounding boxes of \textcolor{qred}{\ours{}}, \textcolor{qblue}{VQLoC~\cite{jiang2023vqloc}}, \textcolor{qgreen}{ground-truth boxes}, and a confidence score curve.
    }
    \figmargin
    \label{fig:partial}
\end{figure*}

%% file: figure/fig_qual_apperance.tex
\definecolor{qgreen}{HTML}{41A24D}
\definecolor{qred}{HTML}{EB321A}
\definecolor{qblue}{HTML}{0025F6}

\begin{figure*}[t]
    \centering
    \includegraphics[width=\linewidth]{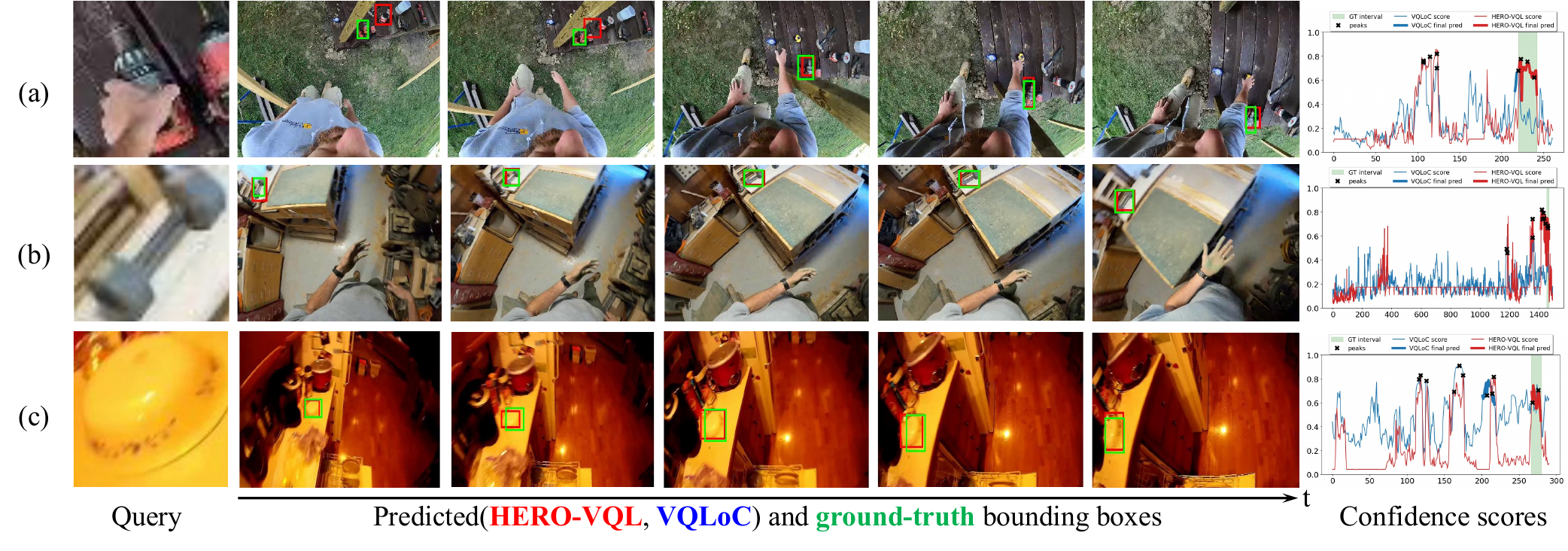}

    \vspace{-0.5em}
    \figcaption{\tb{Qualitative examples in videos with fast-moving object}}
    \figmargin
    \label{fig:appearance}
\end{figure*}

%% file: figure/fig_qual_lemonjin.tex
\definecolor{qgreen}{HTML}{41A24D}
\definecolor{qred}{HTML}{EB321A}
\definecolor{qblue}{HTML}{0025F6}

\begin{figure*}[t]
    \centering
    \includegraphics[width=\linewidth]{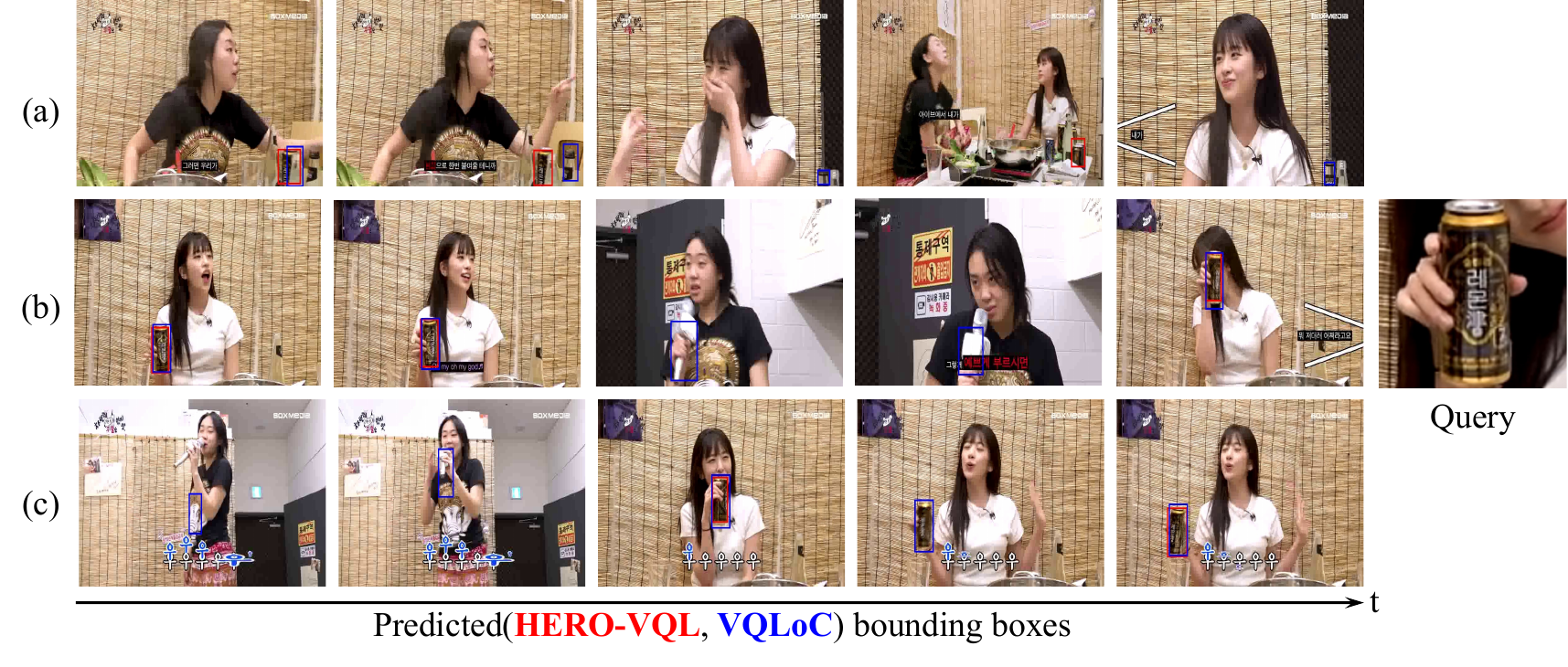}

    \vspace{-0.5em}
    \figcaption{\tb{Qualitative examples in an unseen and real-world video from YouTube}}
    \figmargin
    \label{fig:lemonjin}
\end{figure*}